# One Size Does Not Fit All

## Setting Inference Depth from the Questions a Deployment Actually Asks

Jerry Kaplan[1]
Adjunct Lecturer, Computer Science and Masters in International Policy
Stanford University
September 12, 2026

## Abstract

A transformer language model is trained to respond to any prompt, but each deployment asks only a narrow range of questions: a support assistant sees delivery complaints, a coding tool sees Python. Every deployment nonetheless pays the same computation per token. This paper measures how much of that cost is avoidable when the range of prompts is known in advance.

The mechanism examined is early exit: a small, trained component – called a readout – is attached to an intermediate layer and proposes a token, and a confidence test decides whether to emit it or to run the remaining layers. The models are frozen, and the only supervision used is the model's own output on ordinary traffic.

Three findings are reported. First, achievable savings depend strongly on the kind of traffic: at half depth on a 1.5-billion-parameter model, 96 percent of tokens could be emitted early for arithmetic word problems and 8 percent for Chinese-language explanations, at matched token-level fidelity to the full model (a measure whose limits the third finding exposes). Second, studies of three possible ways a deployment might use knowledge of its traffic found that only customizing the threshold for exiting early is worthwhile: calibrating it per deployment raised exit rates by up to 59 percentage points across three models, and by more than 10 points on most corpora tested, while other techniques showed far less impact. Third, token-level fidelity – the standard evaluation measure in early-exit literature – fails catastrophically in domains where some tokens can be evaluated against ground-truth: on arithmetic word problems, three models each answered sixty questions correctly when run in full, and between 10 and 28 correctly under early exit, in the configuration that scored highest on fidelity.

The intended setting is small models on personal devices. On such hardware, generation is typically limited by memory bandwidth rather than by computation; skipping layers saves both, and early exit is one of several methods that compete in that setting.

## 1. Introduction

Producing one token of output from a transformer language model usually requires a complete pass through the network: every layer, for every token. A layer is one repeated block of arithmetic; a 28-layer model runs 28 of them per token. The word "the" in the middle of a stock phrase takes exactly as much computation to produce as the final digit of a calculation.

A body of prior work observes that this uniformity is wasteful and proposes exiting the computation early when the next token appears determined before the last layer (Schuster et al., 2022; Elhoushi et al., 2024; Shen and Zou, 2026). Two obstacles have kept such methods from being straightforward. First, exiting early requires deciding when to exit, and the decision itself costs computation: judging whether the network has settled on a token requires predicting what token it would produce, which entails a large matrix multiplication against the vocabulary. The second is bookkeeping. Later tokens attend to the internal state of earlier ones at every layer, so a token produced without running the upper layers leaves those layers' state unrecorded, and something must be supplied in its place; Shen and Zou (2026) name this the KV Cache Absence problem.

[1]Corresponding author: Jerry Kaplan, Adjunct Lecturer, Computer Science and Masters in International Policy, Stanford University. (JerryKaplan@Stanford.edu)

This paper does not propose a new exit mechanism; instead it proposes a process for designing and implementing efficient early-exit procedures. We measure how much it matters to know what sort of prompts the deployment will actually receive, and explain how to put this knowledge to productive use. We draw these results from seven kinds of traffic across three models.

All code, prompt sets, and result files needed to reproduce these measurements are available at https://github.com/jerrykaplan/question-conditioned-early-exit.

### 1.1 Why this class of applications

When a model generates text for a single prompt, the time to produce each token is generally dominated by moving the model's parameters (weights) into the processor's fast memory rather than by the computation performed on them. Data-center deployments – typically of large "frontier" models – escape this by serving many prompts at once, so a single parameter fetch contributes to a token for each of them. Speculative decoding techniques exploit the same principle from a different direction: a small substitute model proposes several tokens for one prompt, and the full model checks them all in a single pass, so one set of parameter fetches can yield several tokens (Leviathan et al., 2023; C. Chen et al., 2023).

Models processing traffic from one user can't share parameter fetch costs, because only one prompt is processed at a time, and its tokens must be produced in sequence, since each token depends on the ones before. Speculative decoding applies in this setting as well: the verification pass reads the parameters once for several proposed tokens, and speedups of roughly two-fold on single-user hardware have been reported for drafts produced by a 4-bit copy of the target model (Georganas et al., 2025), and are routinely obtained with smaller models of the same family. Early exit takes a complementary route – doing less work per token – and its savings apply to computation and parameter fetch alike. The two are not exclusive, and the traffic dependence measured in this paper is expected to bear on drafting policies as well, since both accept cheap tokens where they are reliable.

Because skipping higher layers eliminates both the associated computation and parameter fetch costs, and both scale together when layers are skipped, we measure savings in "layer-equivalent" units throughout this paper.

Models in the 1- to 8-billion-parameter range fit adequately on today's consumer devices ("edge computers"), and industry projections for 2026 anticipate that a majority of new personal computers will include dedicated neural processing hardware (GPUs). The economic argument for running such models locally is latency, privacy, and per-query cost.

Within this setting, a deployment that anticipates what it will be asked can be configured to require substantially less computation per token, and this form of optimization will become increasingly useful as such deployments become increasingly common. Note that this does not require training specialized models for each deployment, nor does it restrict what class of questions can be

answered – it only changes how efficiently it produces those answers in a given setting: The same pre-trained model might be used for coding in one deployment, while for writing text, answering medical questions, or single-language prompts (i.e. English only or Mandarin only) in others.

## 2. Definitions of Terms

The following terms are used throughout. (Readers familiar with transformer internals may skip this section.)

*Token.* The unit of text the model reads and writes, typically a short word or a fragment of a longer one. A model with a vocabulary of approximately 150,000 tokens chooses one of those tokens at each step.

*Layer.* One repeated block of arithmetic within the network. Models in this paper have between 24 and 32 layers. Producing one token normally runs all layers in order.

*Residual stream.* The standard term for the running vector of numbers that each layer reads from and adds to. It is the network's working state for the token currently being produced, and has one entry per model dimension: 1,536 for the 1.5-billion-parameter model used here, 3,584 for the 7-billion-parameter model. Because each layer adds to this vector rather than replacing it, the stream at any depth holds the superimposed contributions of every layer below, and information is generally carried by directions through the space rather than by individual entries. Where this paper refers to a layer's "state," it means the residual stream at that layer.

*Unembedding.* The final matrix multiplication that converts the residual stream into a score for every token in the vocabulary. For the 1.5-billion-parameter model used here, this single operation costs about as much computation as five ordinary layers.

*Readout, or lens.* A small, trained component, in this paper a single square matrix plus an offset, that converts the residual stream at an intermediate layer into the coordinate system the model's top layer (head) unembedding expects (Belrose et al., 2023). Without it, intermediate states are largely unreadable by the unembedding; with it, they are far more interpretable. As a concrete illustration of scale, the matrix is approximately three megabytes for a 1.5-billion-parameter model. The readout does not replace the unembedding: its output is passed through the model's own final normalization and unembedding, unchanged.

*Early exit.* Producing a token from an intermediate layer using a readout, rather than running the remaining layers.

*Exit depth.* The layer at which the readout is attached, reported both as a layer number and as a fraction of the model's total layers, because the two are not comparable across models of different depths.

*Confidence gate.* A rule that decides, for each token individually, whether to accept the early answer or to run the remaining layers instead. The rule used here accepts the early answer when the readout's probability for its best token exceeds a threshold.

*Threshold.* The number the confidence gate compares against. A high threshold accepts few early answers and is therefore safe and slow; a low threshold accepts many and is fast and risky. Choosing this number well, per deployment, is the central subject of this paper.

*Exit rate.* The fraction of tokens produced from the intermediate layer rather than from the full network. The higher the exit rate, the more computation is skipped.

*Key-value cache.* The stored internal state of previously produced tokens, which later tokens consult. A token produced by early exit leaves this state incomplete at the layers that were skipped, requiring some proxy for the missing values.

*Nucleus.* The set of most probable tokens that together account for a fixed share of probability, here 90 percent (Holtzman et al., 2020). Deployed systems ordinarily sample from this set rather than always taking the single most probable token, to produce more variety and more natural-sounding responses (but at a cost in overall quality).

*Fidelity.* Whether the token predicted by early exit matches the actual token that would have been selected if all layers of the model were run. Note that the predicted tokens can be correct when the actual full-model token is inappropriate or "wrong", so this is not the same as comparing to "ground truth", when such information is available.

*Acceptance.* Our token-level fidelity measure: an early answer is counted acceptable if it falls within what would have been the full model's nucleus at that position. Section 9 examines the limits of this measure.

*Layer-equivalent.* Our unit of computational cost: the arithmetic of one ordinary layer of the model in question. Costs are reported in this unit so that they do not depend on the particular hardware used.

## 3. The Corpora

Seven bodies of prompts were constructed, sixty prompts each. They were chosen to vary along identifiable axes rather than to be merely different from one another, and two of them exist specifically to isolate one axis while holding content constant. All prompts and the code that generated them are provided in Appendix A and in the accompanying repository.

### 3.1 Explanatory prose (the reference corpus)

Sixty open-ended requests for explanation of everyday phenomena, in the form "Explain why the seasons change" or "Describe how a bill becomes a law in the United States." No system prompt, no imposed format. This corpus stands for the general-purpose assistant traffic against which the restricted corpora are compared.

### 3.2 Python function completion

Sixty requests to complete a named Python function given its signature and a one-line description. Each prompt consists of a single instruction line followed by the function header and docstring, as in the following complete example:

```
Complete this Python function:
def binary_search(arr, target):
    """Return the index of target in sorted arr, or -1."""
```

This corpus is included because code is syntactically constrained in a way prose is not, and because coding assistants are a common deployment category.

### 3.3 Templated information extraction

A single fixed instruction — extract the key facts as a bulleted list of at most five statements — applied to sixty short passages in a uniform news register. This corpus represents the highly restricted end of the range: fixed instruction, fixed output format, and input text drawn from a narrow band of styles. It is the closest available analog to a document-processing feature embedded in a product.

### 3.4 Chinese-language explanation (language held against content)

The same sixty topics as the explanatory prose corpus, written in Chinese. Because the subject matter is identical and only the language differs, any difference in results isolates the effect of language. Section 6 reports what was measured.

### 3.5 Arithmetic word problems

Sixty short quantitative problems with known correct answers, in the form "A car travels 75 miles using 5 gallons. What is the mileage per gallon? Show your work." This corpus is unlike the other six in one respect that proved decisive: its answers can be graded. A second, disjoint set of sixty problems of the same form was generated for the accuracy measurements of Section 9, so that those measurements use problems no component was fitted on.

### 3.6 Customer-service replies

A single fixed instruction establishing the role of a support agent for an electronics retailer, applied to sixty distinct customer complaints ("My order arrived with a cracked screen"). This is a commercially realistic instance of a deployment with a fixed system prompt.

### 3.7 Explanatory prose under a fixed instruction (format held against content)

The same sixty prompts as the explanatory prose corpus, each preceded by a fixed instruction specifying audience, format, and length. Content is identical to Section 3.1 and only the presence of the fixed instruction differs, so any difference isolates the effect of constraining the output format rather than the subject matter.

## 4. Method

The procedure has five steps, none of which modifies the model or requires labeled data.

*Step one: collect traffic.* A minimum of sixty prompts of the kind the deployment is expected to receive are run through the unmodified model, and the internal state of every layer is recorded at every token produced. The model's own choice of token at each position serves as the training target for what follows; no manual annotation is involved.

*Step two: fit a readout at each of some selection of potential layers*, one of which will be selected in the next step. (In our tests, the candidate layers were at approximately one third, one half, and two thirds of the network's depth.) A single square matrix and offset are fitted at each candidate intermediate layer, trained to reproduce the full model's own token choice from the intermediate state. The fitted map is applied to the residual stream at that layer, and the result is passed through the model's own final normalization and unembedding. Fitting details are given in Appendix B.

*Step three: choose the exit depth.* Select the layer whose cost and acceptance best suit the deployment. This choice is made once, in advance of deployment. More than one exit point is possible in principle but rarely economic, because each additional confidence test costs a full unembedding, which can rival the savings from the early exit. (Section 8 expresses this cost in “layer-equivalents”, depending on model size.)

*Step four: calibrate the threshold.* For the specific deployment's traffic, the readout's confidence is computed at every token of a held-out sample, along with whether its choice would have been acceptable (appears within the full model nucleus). From this, the lowest threshold is selected at which the required token-level fidelity is met — for example, that 90 percent of early answers fall within the full model's nucleus. This is a one-time offline computation producing a single number. In an actual deployment, the user would set this threshold depending on application requirements.

*Step five: run.* In deployment, each token runs the layers up to the exit depth; the readout produces a candidate token and a confidence; if the confidence exceeds the calibrated threshold, the token is emitted, and otherwise the remaining layers are run and the full model's answer is used. Tokens produced early leave the upper-layer cache incomplete, and the missing entries are supplied by the established method of projecting the exit-layer state through the upper layers' key and value matrices (Elbayad et al., 2020; Schuster et al., 2022). Shen and Zou (2026) describe a more recent alternative in which the missing entries are generated as a byproduct of the exit computation itself.

## 5. What Parameters are Worth Specializing for a Given Deployment

There are three potential parameters a deployment might set from knowledge of its own traffic: the confidence threshold, the readout's fitted weights, and the exit depth. This section examines them separately, in order of how useful our results suggest they can be.

### 5.1 The confidence threshold: worth up to 59 percentage points of exit rate

The central measurement was made on Qwen2.5-1.5B-Instruct, a 28-layer model, with the readout attached at layer 14 — exactly half of the network's depth. The fidelity requirement was that 90 percent of early answers fall within the full model's nucleus. Two configurations are compared. In the first, a single threshold is used for all traffic, set at the lowest value that meets the fidelity requirement on every one of the seven corpora; this is what a deployment must do when it does not know what it will be asked. In the second, the threshold is calibrated separately for each corpus, which is what a deployment can do when it does know.

| Corpus | Calibrated threshold | Exit rate, single global threshold | Exit rate, calibrated threshold | Gain |
|---|---|---|---|---|
| Arithmetic word problems | 0.30 | 50.4% | 96.4% | +46.0 points |
| Templated extraction | 0.62 | 30.0% | 66.9% | +36.9 points |
| Customer-service replies | 0.70 | 21.8% | 56.8% | +34.9 points |
| Explanatory prose | 0.89 | 9.1% | 21.8% | +12.7 points |
| Python completion | 0.93 | 18.2% | 29.3% | +11.1 points |
| Prose, fixed instruction | 0.91 | 8.4% | 19.2% | +10.8 points |
| Chinese explanation | 0.98 | 7.6% | 7.6% | +0.0 points |

*Table 1. Fraction of tokens produced without running the full network, matched token-level fidelity, on a 28-layer model with the readout at layer 14. The single global threshold is 0.98, set by the hardest corpus.*

The single safe threshold is 0.98, and it is set by Chinese-language traffic, the most demanding of the seven. Every other kind of traffic is then run more conservatively than its own measurements would require. Arithmetic traffic, which can safely produce 96 percent of its tokens from half the network by exiting early, exits 50 percent of the time instead. Extraction traffic, which can safely exit 67 percent of the time, exits only 30 percent of the time when held to the more stringent single-global-threshold standard.

The mechanism is visible in the first column. The threshold that meets the fidelity requirement varies from 0.30 to 0.98 across these corpora — more than a factor of three — and a system serving unknown traffic has no way to use anything but the largest of them. Chinese traffic gains nothing from calibration for the same reason: it is the constraint. The same comparison was run on Qwen2.5-7B-Instruct and on Llama-3.1-8B-Instruct, with gains of up to 59 points and up to 50 points respectively, and with Chinese setting the global threshold in both. The gain is zero for any corpus whose own calibrated threshold already equals the globally safe one, which on the 7-billion-parameter model applies to English prose and Chinese alike. The effect is therefore not particular to one model or one model family. Complete per-model tables are given in Appendix D.

### 5.2 The readout's weights: worth about two percentage points

The readout's weights are the contents of the matrix fitted at a given layer; a readout "fitted to prose" and a readout "fitted to code" may sit at the same layer and differ only in their coefficients. The question addressed here is whether the readout's weights repay specialization by corpora, which is a proxy here for different deployments. All measurements in this section were made on Qwen2.5-0.5B-Instruct, a 24-layer model, with every readout attached at layer 12 — the location is held fixed throughout, so that only the weights vary.

| Readout weights fitted on | Tested on prose | Tested on code | Tested on extraction |
|---|---|---|---|
| Prose only | 0.663 | 0.165 | 0.279 |
| Code only | 0.419 | 0.622 | 0.261 |
| Extraction only | 0.177 | 0.072 | 0.773 |
| All three combined | 0.635 | 0.606 | 0.779 |

*Table 2. Acceptance rates for readouts fitted to single corpora and to all corpora combined, all attached at layer 12 of a 24-layer model. Each row is one fitted matrix; each column is one test corpus. The combined readout was fitted on three times as much data, so that its advantage in sample size does not confound the comparison. Fitting details are in Appendix B.*

Reading down the columns, the readout fitted to all three corpora is as good as each specialist on that specialist's own traffic, within about two percentage points.

Reading across the rows shows why this outcome is not evidence that the corpora are similar. A readout fitted on prose scores 0.663 on prose and 0.165 on code — it is nearly useless on traffic it was not fitted to. The internal states really are strongly specific to the kind of traffic. The combined readout succeeds because a matrix of roughly 800,000 parameters (as in our test) has enough capacity to represent all three specializations at once, not because there is only one specialization to represent. One readout therefore serves all seven corpora, which means it can be fitted once by the model's publisher rather than separately by each deployment, and a deployment that sees mixed traffic need not detect the type and switch between readouts.

### 5.3 The exit depth: go as shallow as the fidelity requirement permits

The exit depth is a choice of location for the readout. Deeper readouts read more accurately, because more of the network's computation has been performed; shallower readouts save more, because more layers are skipped. Since both effects are monotonic, the question is whether they trade off against one another to produce an interior optimum. In the range we measured, they do not.

| Model | Exit depth | Acceptance | Exit rate at threshold 0.85 | Estimated speedup |
|---|---|---|---|---|
| Qwen2.5-1.5B (28 layers) | 9 (32%) | 0.588 | 35.8% | 1.26 |
| | 14 (50%) | 0.631 | 40.9% | 1.21 |
| | 18 (64%) | 0.696 | 46.4% | 1.16 |
| Qwen2.5-7B (28 layers) | 9 (32%) | 0.563 | 44.1% | 1.38 |
| | 14 (50%) | 0.596 | 45.5% | 1.27 |
| | 18 (64%) | 0.636 | 49.5% | 1.20 |
| Llama-3.1-8B (32 layers) | 10 (31%) | 0.618 | 48.5% | 1.45 |
| | 16 (50%) | 0.687 | 54.9% | 1.34 |
| | 21 (66%) | 0.787 | 63.8% | 1.26 |

*Table 3. Acceptance and estimated speedup by exit depth, pooled across all seven corpora. Acceptance rises with depth in every model; estimated speedup falls monotonically, because the layers saved shrink faster than the exit rate grows.*

The practical guidance is therefore simple: place the readout as shallow as the deployment's fidelity requirement permits, and let that requirement rather than a cost optimum determine the depth. A deployment with a demanding fidelity bar will be pushed deeper and will save less; a deployment with a tolerant one can sit near a third of the network's depth. We did not test depths below roughly 30 percent, where acceptance would eventually fall far enough that the fallback path runs on most tokens and the savings disappear; that boundary has not been located.

## 6. What the Seven Corpora Show

Two observations follow from Table 1 and the acceptance measurements underlying it.

**Language matters more than any other axis tested.** Chinese explanation had the lowest acceptance rate given exit (0.69 against 0.79 for the identical content in English) and the highest calibrated threshold of the seven corpora (0.98). Because the topics are identical to the English prose corpus and only the language differs, this is attributable to language rather than to subject matter. The effect appeared in all three models, in both families tested. Wendler et al. (2024) report that multilingual transformers pass through distinct phases as computation proceeds through the layers, with language-specific output produced late; our measurement is consistent with that account, though we did not test the mechanism directly.

**Constraining the output format has a real but small effect.** Fixed-instruction prose (requesting a concise, bulleted response) showed slightly higher acceptance than the same content without the instruction (0.543 against 0.523) and a slightly higher calibrated threshold. The large differences in Table 1 are therefore attributable mainly to the kind of content and its intrinsic predictability, not to the mere presence of a fixed instruction.

## 7. Result: Early Exit Requires a Fallback Path

It is natural to ask whether the confidence gate is necessary at all. If calibration establishes that a given kind of traffic can be produced acceptably from layer 14, one might simply produce every token from layer 14, which would remove the cost of computing confidence and the need to repair the cache. This was tested and does not work.

On a 24-layer model with the readout at layer 12, producing every token from the intermediate layer yielded a perplexity of 178 against 1.70 for the unmodified model, with text remaining coherent for a median of about eight tokens before noticeably degenerating. Producing every token at a fixed schedule — running the full network every second or fourth token regardless of confidence — was also insufficient, giving a perplexity of 14.4 at half the full cost, while the confidence gate at the same cost gave 3.2.

The cause is a feedback effect that token-level measurements cannot reveal. An acceptance rate of 0.7 means that roughly three tokens in ten fall outside the full model's nucleus. Each such token then becomes part of the context for the next token. The readout was fitted on text the full model produced, so it is least reliable on text that has begun to drift, and the drift compounds.

This diagnosis was tested directly. The readout was refitted on 2,494 token positions taken from the early-exit system's own degenerate output, labeled with the full model's judgment at those same positions — a correction that requires no human labels and can be performed on deployment traffic, following the approach of Ross et al. (2011) for correcting distribution shift in sequential decision problems. The refitted readout improved perplexity from 178 to 16.0, an elevenfold improvement that confirms the diagnosis, and left coherent output length essentially unchanged at nine tokens. The failure is therefore not merely a mismatch between fitting and deployment conditions; it is structural, and follows from the compounding arithmetic of per-token acceptance.

Restricted traffic results in a higher exit rate, but not immunity from this failure. On the 7-billion-parameter model, templated extraction generated at a threshold of 0.85 produced text of perplexity 4.23, while the same corpus at a threshold of 0.7 collapsed to perplexity 17.34 with coherent output lasting a median of eight tokens. The corpus that tolerates the most early exit therefore also has a narrow margin above the point at which generation degrades, which is an argument for calibrating conservatively.

This result explains a design feature shared by most working systems in the literature. The fallback to the full network is the mechanism that keeps generated text anchored, and it is why reported speedups for lossy early-exit methods are modest. How much accumulated drift a full-model step can absorb before the text is beyond recovery is discussed in Section 11. Note that allowing a full pass through the model does not completely reset the context, because the tokens already emitted remain in the residual stream.

## 8. Closed-Loop Measurements and Cost Accounting

Table 4 reports generated text, not token-level statistics: each corpus was generated with the confidence gate at layer 14 of 28 with a threshold of 0.85, and the resulting text was scored by the unmodified model (as measured by perplexity). The measurements are from Qwen2.5-7B-Instruct. Perplexity is used here as a screen for gross degeneration — the failure mode of Section 7, where generation collapses into repetition — and not as evidence that quality is preserved. Section 9 shows why it cannot serve the latter purpose.

| Corpus | Exit rate | Perplexity | Speedup |
|---|---|---|---|
| Arithmetic word problems | 60.2% | 2.53 | 1.38 |
| Templated extraction | 51.7% | 4.23 | 1.31 |
| Customer-service replies | 50.9% | 2.99 | 1.31 |
| Python completion | 41.6% | 3.70 | 1.24 |

| Chinese explanation | 26.8% | 3.64 | 1.14 |
|---|---|---|---|
| Explanatory prose | 26.0% | 2.87 | 1.14 |
| Prose, fixed instruction | 24.3% | 3.61 | 1.13 |

*Table 4. Generated-text perplexity measurements on the 7-billion-parameter model, 28 layers, readout at layer 14, threshold 0.85. The unmodified model's perplexity on its own output is 1.33. Speedups are estimates derived from computation counts.*

The ordering matches the offline calibration: traffic that calibration identifies as permitting more early exits does in fact run faster in generation, by a factor ranging from 1.13 to 1.38 on this model. On the 8-billion-parameter Llama model the same procedure gives speedups of 1.20 to 1.59, with perplexities between 1.67 and 2.47 against a baseline perplexity of 1.30 on the unmodified model, so the pattern holds across model families. The arithmetic row must be read together with Section 9, which shows that perplexity is a poor guide to quality on traffic where individual tokens can be checked against a known correct answer.

Speedup in Table 4 is calculated by estimating the amount of computation saved by skipping the layers above the readout layer, minus the cost of performing the early-exit tests. Most of this cost is the confidence decision itself, which requires running the unembedding. That cost depends strongly on model size, because the vocabulary is fixed while layers grow wider: it is 9.2 layer-equivalents for the 0.5-billion-parameter model, 5.0 for the 1.5-billion, 2.4 for the 7-billion, and 2.5 for the 8-billion Llama model. Measured against stack depth, the same decision costs 38 percent of the network at the smallest size and 8.5 percent at 7 billion parameters. The derivation is given in Appendix C.

These figures predict elapsed time from exact operation counts, so they were checked against a clock. Running the unmodified 7-billion-parameter model and the gated system on an Apple M5 MacBook Air laptop with 32 gigabytes of unified memory, at the exit rates of Table 4, gave measured speedups of 1.12, 1.30 and 1.35 for exit rates of 26, 52 and 60 percent, against estimates of 1.14, 1.31 and 1.38 — within two percent in each case, and slightly below the estimate throughout, as would be expected of a model that omits implementation overheads. In absolute terms the unmodified model produced 7.5 tokens per second on that machine and the gated system 10.1. Details are given in Appendix C.

This is visible in the measurements. At a threshold of 0.85, the same corpora that yield speedups of 1.04 to 1.25 on the 1.5-billion-parameter model yield 1.13 to 1.38 on the 7-billion-parameter model. Speedup figures measured on small models therefore understate what the same method achieves at deployment scale, and this is one reason the method is more valuable at models of a few billion parameters rather than smaller ones.

## 9. Result: Token-Level Fidelity Does Not Measure Task Correctness

The measure used above, and in the literature generally, asks how often the early answer would have been acceptable to the full model at that position. It is computed token by token and averaged uniformly. Arithmetic word problems make the negative consequence of that uniformity visible, because their numerical answers can be graded against a known correct value.

Sixty arithmetic problems, disjoint from those used to fit any readout, were run through each of the three models unmodified and through the same models with a confidence gate at half depth.

| Model | Unmodified | Gate, threshold 0.7 | Gate, threshold 0.85 |
|---|---|---|---|
| Qwen2.5-1.5B (28 layers) | 60 / 60 | 10 / 60 (exit 55%) | 11 / 60 (exit 46%) |
| Qwen2.5-7B (28 layers) | 60 / 60 | 12 / 60 (exit 69%) | 34 / 60 (exit 57%) |
| Llama-3.1-8B (32 layers) | 60 / 60 | 28 / 60 (exit 78%) | 40 / 60 (exit 74%) |

*Table 5. Arithmetic problems answered correctly, out of sixty, with the fraction of tokens emitted by early exit in parentheses. All three models answer every problem correctly when run in full.*

Every model tested answered all sixty problems correctly when run in full. At a threshold of 0.7, where between 55 and 78 percent of tokens are emitted early, between half and five sixths of that competence is lost; at 0.85 the loss ranges from a third to four fifths. The threshold governs how much is lost, but not reliably: raising it from 0.7 to 0.85 recovers a large share on the 7-billion-parameter model (12 correct goes to 34 correct) and almost none on the 1.5-billion-parameter model (10 correct only goes to 11). A practitioner cannot assume that a more conservative threshold buys correctness back.

This is an important distinction because no measure in ordinary use reports it in the published literature. The same 7-billion-parameter configuration that answers twelve of sixty problems correctly at a threshold of 0.7 has a token-level acceptance rate of 0.874 — the highest of all seven corpora — and generates fluent text of perplexity 3.55 against a baseline of 1.33.

The failure is visible in the text. Given "A car travels 75 miles using 5 gallons. What is the mileage per gallon?", the unmodified model divides correctly and reports 15 miles per gallon. The gated system writes a correct-looking framing — "To find the mileage per unit length, you need to divide the total sum of the units of measurement by the number of units" — then states "Miles: 1000", a figure that appears nowhere in the prompt, and in subsequent text, degenerates into repetition.

The reason the fidelity measure does not detect this is that the response is roughly two hundred tokens, of which perhaps five carry the answer. An error confined to those five tokens moves an average over two hundred tokens very little. Perplexity behaves the same way. The acceptance measure is additionally permissive at exactly those positions, because immediately before a digit is produced many different digits are individually plausible continuations.

Arithmetic is not a special case; it is the case where the error is detectable. The same substitution in explanatory prose — a wrong date, a wrong name, a reversed comparison — would produce the same fidelity scores and would not be caught by any measure in common use, because such text has no gradable answer. Therefore we recommend that evaluations of early-exit systems include at least one corpus with checkable answers, reported separately, and that token-level agreement not be treated as sufficient evidence of preserved quality. This applies to the present paper as well. The six corpora without gradable answers are evaluated here only by token-level acceptance and by perplexity, so nothing reported above establishes that their perceived quality is preserved at the operating points shown; the exit rates are measurements, the fidelity figures are a screen, and the question of what quality those configurations actually cost is taken up in the work described in Section 11.

A final observation concerns which traffic is most exposed. Arithmetic scored the highest acceptance of the seven corpora because its supporting prose is formulaic and easy to predict early. That property is logically independent of the fact that the correctness of its answers hinges on a few tokens, but the two occur together in this corpus, and their co-occurrence is what makes the configuration hazardous: the traffic that a fidelity measure rates as safest to exit early is also the traffic where a single wrong token ruins the response, in this case. Structured or templated output is therefore not an escape from this problem; indeed it may be where the problem is most acute.

## 10. What the Readout Does, and Why Nothing Simpler or More Elaborate Serves

The readout is the one fitted component in the method, and its form was not chosen arbitrarily. Several alternatives were measured, all on Qwen2.5-0.5B-Instruct, at layer 12 of the 24-layer model, all evaluated on held-out answers.

Applying the model's final (top layer) unembedding to the intermediate residual stream directly, with no transformation fitting it to what the unembedding expects, gives an acceptance of 0.003 to 0.014 depending on corpus — effectively nothing. The information is present at that layer but not in coordinates the unembedding can read.

We tested various alternatives to using a full matrix to transform the residual stream into the unembedding's coordinate system, to see if any of this exit cost could be saved. Constraining the transformation to a pure rotation, fitted in closed form by the standard orthogonal Procrustes solution (Schönemann, 1966), raises acceptance only to 0.05 to 0.10. Adding a single global scale factor, which corrects for the fact that intermediate residual streams are smaller in magnitude than final-layer ones (the fitted factor is 10.4 for this model and layer), raises it much further, to 0.35 to 0.47. Using an unconstrained fitted matrix reaches 0.61 to 0.78. Examination of its singular values explains the remaining gap: the median singular value is 1.01, so the map leaves a typical direction unchanged in magnitude, but the ratio of largest to smallest is 118, meaning a minority of directions are suppressed by up to fiftyfold. That selective suppression is what a rotation and a uniform scale cannot express.

Replacing the matrix with a small neural network — the same affine map plus a residual branch with one hidden layer of 896 units and a nonlinearity, tripling the parameter count — raises held-out acceptance from 0.638 to 0.645, while training accuracy rises to 0.954. The added capacity is spent memorizing training positions and does not result in a materially better acceptance rate. The limitation is therefore not the expressiveness of the readout but the information available at that depth.

Two cheaper substitutes were also tested for the confidence test, since the unembedding it requires is the dominant cost. The first was comparing residual streams at adjacent layers, on the theory that a residual stream which has stopped changing has likely already settled on which tokens it is considering emitting. This approach produces a signal whose rank correlation with acceptance is between −0.12 and +0.11, and which selects tokens no better than chance at matched exit rates; first removing the corpus-wide mean residual at this level does not improve it.

We also tried fitting a logistic classifier directly on the residual coordinates. This does better but not by much: reading all 896 coordinates gives an area under the curve of 0.59 to 0.75, against 0.78 to 0.85 for the unembedding-based confidence. On this evidence, the information about whether an exit is safe is not linearly available in the residual coordinates and becomes available only after projection into vocabulary space, though we do not claim that no cheaper sufficient statistic exists.

Finally, the choice among softmax-based confidence measures is not consequential. The probability of the best token, the margin between the best and second-best, and the negative entropy of the readout's distribution all deliver acceptance within about two percentage points of one another at matched exit rates, and all three require the same full projection. Acceptance is monotone in confidence in every corpus, rising from 0.32 in the lowest decile to 0.99 in the highest on arithmetic and from 0.06 to 0.92 on prose.

## 11. Limitations

- Speedups reported in the main tables are predictions of elapsed time, obtained by counting the operations and parameter fetches saved. The counts themselves are exact; what is estimated is the time they translate into, since real hardware carries overheads the counts omit. Section 8 reports a measured comparison on a consumer laptop that agrees with the predictions to within two percent, at exit rates imposed to match the measured ones rather than computed from confidence at run time.
- The generated-text measurements use twelve held-out prompts per corpus. The differences between corpora are large relative to that sample, but the individual figures are not precise. The arithmetic corpus accuracy measurements of Section 9 use sixty problems per condition.
- Quality is assessed by token-level acceptance, by perplexity under the unmodified model, and — for arithmetic alone — by answer correctness. Section 9 shows that the first two can certify a configuration that has lost most of the model's competence, so for the five corpora with no gradable answer this paper reports no measurement that establishes preserved quality. Subsequent work currently in preparation by the author addresses this directly, using a calibrated judge with null controls and stated resolution to measure the quality cost of early exit and other lossy optimizations on these same models and corpora; that work finds the losses on open-ended traffic to be real and domain-dependent, and confirms that fidelity at the token level does not predict them.
- The Chinese corpus was translated by the authors without a native-speaker review, so a translation artifact cannot be entirely excluded as a contributor to that corpus's results.
- The confidence gate implemented here follows the established design in the literature and has not been optimized. In particular, the cheap confidence signals examined in Section 10 do not exhaust the possibilities: a method that bounds the largest token score without computing all of them would change the cost accounting substantially, and has not been tested.
- How much accumulated drift the full model can absorb before generated text is beyond recovery was not characterized in the experiments reported here. Subsequent measurements by the author, to be reported separately, indicate that when every token exits, error in the reconstructed cache compounds over long responses – a judged quality loss of roughly one rating point at half depth, rising to severe degradation at one third of the depth – and that a periodic full-depth pass over the most recent tokens, which recomputes their cache entries exactly at a cost of about one token-time on bandwidth-bound hardware, recovers much of the loss. The confidence gate's full-depth tokens play the same anchoring role. A related periodic-verification design appears in SpecPV (2025). This bears directly on how conservatively a deployment must be calibrated.
- Results are reported for three models across two families, at 1.5, 7, and 8 billion parameters, with supporting mechanism measurements on a fourth model of 0.5 billion parameters. Generality beyond these architectures is not established.

## 12. Relation to Prior Work

Confidence-gated early exit was established for encoder-decoder models by Schuster et al. (2022), who introduced per-token confidence measurement with a calibrated threshold. LayerSkip (Elhoushi et al., 2024) applies early exit to decoder-only models but requires retraining the base model with layer dropout. River-LLM (Shen and Zou, 2026) provides a training-free token-level exit whose cache repair is produced as a byproduct of the exit computation, and whose exit signal does not depend on the deployment's traffic; it reports practical speedups of 1.71 to 2.16 on mathematical reasoning and code generation. The readout component follows the tuned lens of Belrose et al. (2023). Speculative decoding (Leviathan et al., 2023; C. Chen et al., 2023) occupies the lossless regime and applies to single-user devices as well: drafts produced by a 4-bit copy of the target model (Zhao et al., 2025; Tiwari et al., 2025; Georganas et al., 2025) or by a smaller model of the same family yield roughly two-fold speedups on bandwidth-bound hardware. Kangaroo (Liu et al., 2024) and Draft & Verify (Zhang et al., 2024) draft from shallow layers of the same model – the former through a small adapter with a confidence-based stop to drafting – and are the closest self-speculative relatives of the gate examined here. Per-deployment threshold adaptation also has precedent: Bae et al. (2023) estimate the exit threshold for new data from the shallow model's agreement with the deep one, and an empirical study of early exit in frozen models without joint optimization (Shan et al., 2024) examines the setting assumed here. Relaxed acceptance rules that trade exactness for speed (Cai et al., 2024) and utility-based acceptance that rejects only tokens which change the final answer (Ziashahabi et al., 2025) reach the conclusion of Section 9 from the other direction: distributional fidelity is neither necessary nor sufficient for task correctness. A framework for training and serving early-exit models at scale, including two inference approaches compatible with KV caching, is given by Y. Chen et al. (2024).

Relative to this body of work, the contribution of the present paper is not a mechanism. It is the measurement of how strongly the achievable operating point depends on the deployment's traffic, the size of the gain available from calibrating the threshold per deployment, using a label-free procedure in the tradition of Schuster et al. (2022) and Bae et al. (2023), and the demonstration that token-level fidelity measures can certify configurations that destroy task correctness.

## 13. Acknowledgments and Declaration of Generative AI and AI-assisted Technologies in the Manuscript Preparation Process

Experiments, analysis, and drafting were carried out in collaboration with Claude Fable 5.1 and Claude Opus 5 (Anthropic), AI systems operating under the author's direction across an extended series of working sessions. All experimental designs were approved and/or originated by the author; all claims were verified against archived experimental records. Responsibility for all content of the published article rests solely with the author.

## 14. References

Bae, S., Ko, J., Song, H., and Yun, S.-Y. (2023). Fast and Robust Early-Exiting Framework for Autoregressive Language Models with Synchronized Parallel Decoding. Empirical Methods in Natural Language Processing. arXiv:2310.05424.

Belrose, N., Furman, Z., Smith, L., Halawi, D., Ostrovsky, I., McKinney, L., Biderman, S., and Steinhardt, J. (2023). Eliciting Latent Predictions from Transformers with the Tuned Lens. arXiv:2303.08112.

Cai, T., Li, Y., Geng, Z., Peng, H., Lee, J. D., Chen, D., and Dao, T. (2024). Medusa: Simple LLM Inference Acceleration Framework with Multiple Decoding Heads. arXiv:2401.10774.

Chen, C., Borgeaud, S., Irving, G., Lespiau, J.-B., Sifre, L., and Jumper, J. (2023). Accelerating Large Language Model Decoding with Speculative Sampling. arXiv:2302.01318.

Chen, Y., Pan, X., Li, Y., Ding, B., and Zhou, J. (2024). EE-LLM: Large-Scale Training and Inference of Early-Exit Large Language Models with 3D Parallelism. arXiv:2312.04916, International Conference on Machine Learning (ICML 2024).

Elbayad, M., Gu, J., Grave, E., and Auli, M. (2020). Depth-Adaptive Transformer. International Conference on Learning Representations.

Elhoushi, M., Shrivastava, A., Liskovich, D., Hosmer, B., Wasti, B., Lai, L., Mahmoud, A., Acun, B., Agarwal, S., Roman, A., Aly, A., Chen, B., and Wu, C.-J. (2024). LayerSkip: Enabling Early Exit

Inference and Self-Speculative Decoding. Annual Meeting of the Association for Computational Linguistics.

Georganas, E., Kalamkar, D., Kozlov, A., and Heinecke, A. (2025). ML-SpecQD: Multi-Level Speculative Decoding with Quantized Drafts. arXiv:2503.13565.

Holtzman, A., Buys, J., Du, L., Forbes, M., and Choi, Y. (2020). The Curious Case of Neural Text Degeneration. International Conference on Learning Representations.

Leviathan, Y., Kalman, M., and Matias, Y. (2023). Fast Inference from Transformers via Speculative Decoding. International Conference on Machine Learning.

Liu, F., Tang, Y., Liu, Z., Ni, Y., Han, K., and Wang, Y. (2024). Kangaroo: Lossless Self-Speculative Decoding via Double Early Exiting. arXiv:2404.18911, Advances in Neural Information Processing Systems 37.

Ross, S., Gordon, G., and Bagnell, D. (2011). A Reduction of Imitation Learning and Structured Prediction to No-Regret Online Learning. Artificial Intelligence and Statistics.

Schönemann, P. H. (1966). A generalized solution of the orthogonal Procrustes problem. Psychometrika, 31(1), 1–10.

Schuster, T., Fisch, A., Gupta, J., Dehghani, M., Bahri, D., Tran, V., Tay, Y., and Metzler, D. (2022). Confident Adaptive Language Modeling. Advances in Neural Information Processing Systems 35.

Shan, W., Meng, L., Zheng, T., Luo, Y., Li, B., Wang, J., Xiao, T., and Zhu, J. (2024). Early Exit Is a Natural Capability in Transformer-based Models: An Empirical Study on Early Exit without Joint Optimization. arXiv:2412.01455.

Shen, Y., and Zou, A. (2026). River-LLM: Large Language Model Seamless Exit Based on KV Share. arXiv:2604.18396.

Tan, Z., Zhang, X., Hu, C., Peng, J., and Xia, K. (2025). SpecPV: Improving Self-Speculative Decoding for Long-Context Generation via Partial Verification. arXiv:2512.02337.

Tiwari, R., Xi, H., Tomar, A., Hooper, C., Kim, S., Horton, M., Najibi, M., Mahoney, M. W., Keutzer, K., and Gholami, A. (2025). QuantSpec: Self-Speculative Decoding with Hierarchical Quantized KV Cache. International Conference on Machine Learning. arXiv:2502.10424.

Wendler, C., Veselovsky, V., Monea, G., and West, R. (2024). Do Llamas Work in English? On the Latent Language of Multilingual Transformers. Annual Meeting of the Association for Computational Linguistics, 15366–15394.

Zhang, J., Wang, J., Li, H., Shou, L., Chen, K., Chen, G., and Mehrotra, S. (2024). Draft & Verify: Lossless Large Language Model Acceleration via Self-Speculative Decoding, Proceedings of the 62nd Annual Meeting of the Association for Computational Linguistics (Volume 1: Long Papers), 11263–11282.

Zhao, J., Lu, W., Wang, S., Kong, L., and Wu, C. (2025). QSpec: Speculative Decoding with Complementary Quantization Schemes. Proceedings of the 2025 Conference on Empirical Methods in Natural Language Processing. arXiv:2410.11305.

Ziashahabi, A., Bakman, Y. F., Yaldiz, D. N., El-Khamy, M., Karimireddy, S. P., and Avestimehr, S. (2025). Reject Only Critical Tokens: Pivot-Aware Speculative Decoding. The First Workshop on Efficient Reasoning, NeurIPS 2025. arXiv:2511.00351.

## Appendix A. The Corpora

Each corpus contains sixty prompts. Three were written manually (explanatory prose, Python completion, Chinese explanation) and four were generated from templates (templated extraction, arithmetic, customer-service replies, and fixed-instruction prose). Complete listings and the generating code are in the repository; representative examples and the construction rules are given here.

### A.1 Explanatory prose

Sixty single-sentence requests, no system prompt. The first five are:

```
Explain why the seasons change.
Describe how a bill becomes a law in the United States.
Explain what causes inflation.
Describe the water cycle.
Explain why the Roman Empire declined.
```

### A.2 Python function completion

Sixty prompts of the form shown in Section 3.2, covering string manipulation, search and sorting, numerical routines, and data-structure operations. Function names range from reverse_words and is_palindrome through binary_search and levenshtein to power_set and split_camel.

### A.3 Templated information extraction

One fixed instruction, applied to sixty synthetic passages:

```
Extract the key facts from the following text as a bulleted list of
at most five short factual statements. Use only information in the text.
```

Each passage is assembled from an opener, a middle clause carrying a quantity, and a closer, with a topic drawn from a list of sixty (a city council meeting about a bike lane, a quarterly earnings report, a museum exhibit opening, and so on). A complete example:

```
Text: According to a statement released Tuesday, the item concerning a
new solar farm project has moved forward. Specifically, the project
carries an estimated cost of 7 hundred thousand dollars. A follow-up
meeting is scheduled for later in the spring.
```

### A.4 Chinese-language explanation

The sixty topics of A.1, rendered in Chinese. The first three are 请解释四季变化的原因。 (why the seasons change), 请描述美国的法案如何成为法律。 (how a bill becomes law), and 请解释通货膨胀的成因。 (what causes inflation). The translations were produced by the authors and were not reviewed by a native speaker; see Section 11.

### A.5 Arithmetic word problems

Ten templates instantiated with randomly drawn integers, constrained so that every answer is a whole number. Two sets were generated with different random draws: one used in the corpus comparisons of Sections 5 and 6, and a second, disjoint set of sixty used for the accuracy measurements of Section 9. Examples from the second set, with their answers:

```
A worker earns $75 per hour for 14 hours. What are the total
earnings? Show your work.                                   [1050]
A tank of 77 liters drains at 7 liters per hour. How many hours
until empty? Show your work.                                  [11]
```

Generation used a limit of 400 tokens; no response reached that limit, so no answer was truncated before the calculation was stated. Responses were scored automatically by testing whether the correct integer appeared among the last three numbers in the text.

### A.6 Customer-service replies

One fixed instruction — establishing the role of a support agent for an online electronics retailer, and requiring the reply to acknowledge the issue, state the next step, and give a timeframe — applied to sixty distinct complaints, from “My order arrived with a cracked screen” to “I need proof of delivery for my insurance.”

### A.7 Explanatory prose under a fixed instruction

The sixty prompts of A.1, each preceded by:

```
You are a science explainer for a general audience. Answer in clear,
plain prose. Do not use lists. Keep to one paragraph.
```

## Appendix B. Fitting the Readout

The readout at each candidate layer is a square matrix of the model's width plus an offset vector, initialized at the identity and the zero vector respectively, and fitted by stochastic gradient descent with the Adam optimizer at a learning rate of 0.001 and a batch size of 256, for three passes over the training data. The objective is cross-entropy between the readout's output distribution — obtained by passing its result through the model's own final normalization and unembedding — and the token the unmodified model itself selected at that position. No human-supplied labels are used at any point.

Training data are the recorded residual streams at the chosen layer for every token of every response, split at the level of whole responses: the first 48 of each corpus's 60 responses are used for fitting and the remaining 12 are held out for all reported measurements. Splitting by response rather than by token avoids having positions from the same response appear on both sides of the split; individual token types of course recur across both, in different contexts. Recording used greedy decoding with the repetition penalty disabled, so that the token emitted is exactly the model's highest-scoring token; this was verified by reconstructing the emitted token from the recorded final-layer state, which matched between 98.5 and 99.6 percent of positions, the remainder being ties at the precision of the stored values.

| Corpus (Qwen2.5-0.5B, layer 12) | Training positions | Held-out positions |
|---|---|---|
| Explanatory prose | 9,537 | 2,400 |
| Python completion | 9,600 | 2,400 |
| Templated extraction | 2,025 | 594 |
| All three combined | 21,162 | 5,394 |

*Table B1. Data volumes for the comparison of Table 2. The combined readout was fitted on approximately three times the data of any specialist, so that its advantage in sample size cannot be mistaken for an advantage of generality. The extraction corpus is smaller because its responses terminate naturally after a short bulleted list.*

The negative controls of Section 10 use the same split and the same optimizer settings. The nonlinear variant adds a residual branch with one hidden layer of 896 units and a GELU nonlinearity, with the second weight matrix initialized to zero so that training begins exactly at the fitted affine solution and any change is attributable to the nonlinearity. The rotation-only variant is fitted in closed form rather than by gradient descent, as the orthogonal Procrustes solution of the cross-product between intermediate and final-layer states.

## Appendix C. The Layer-Equivalent Cost Model and Measured Time

One layer-equivalent is the number of multiply-accumulate operations performed by one ordinary transformer layer of the model in question, computed from the model's configuration. For a model of width H with A attention heads, K key-value heads, head dimension D and feed-forward width F, one layer costs $H \cdot A \cdot D + 2 \cdot H \cdot K \cdot D + A \cdot D \cdot H$ for the four attention projections, accounting for grouped-query attention where the key and value projections are narrower, plus $3 \cdot H \cdot F$ for the three matrices of the gated feed-forward block. The confidence test costs H·H for the readout plus H·V for the unembedding, where V is the vocabulary size.

| Model | Width | Vocabulary | Layers | Confidence test (readout + unembedding), in layer-equivalents | Share of full stack |
|---|---|---|---|---|---|
| Qwen2.5-0.5B | 896 | 151,936 | 24 | 9.2 | 38% |
| Qwen2.5-1.5B | 1,536 | 151,936 | 28 | 5.0 | 18% |
| Qwen2.5-7B | 3,584 | 152,064 | 28 | 2.4 | 8.5% |
| Llama-3.1-8B | 4,096 | 128,256 | 32 | 2.5 | 7.8% |

*Table C1. Cost of one confidence test, by model. The test grows cheaper relative to the network as models grow wider, because the vocabulary size is fixed.*

At single-prompt generation the same accounting approximates parameter movement as well as arithmetic, because both scale with the number of parameters touched. For Qwen2.5-7B, one layer holds approximately 255 million parameters and the unembedding approximately 545 million, a ratio of 2.1, against the 2.4 computed from operation counts.

Elapsed time was measured on an Apple M5 MacBook Air laptop with 32 gigabytes of unified memory, running Qwen2.5-7B-Instruct in half precision on the integrated GPU. Four conditions — the unmodified model and the gated system at exit rates of 26, 52 and 60 percent, corresponding to the prose, extraction and arithmetic rows of Table 4 — were run round-robin, twice each after a discarded warm-up round, generating 30 tokens per run, and medians are reported; the repetitions within each condition agreed to within two percent. Exit decisions were imposed at the stated rates rather than computed from confidence, and the readout matrix was randomly initialized, since elapsed time depends on the shape of that matrix and not on its contents; the measurement therefore establishes the cost of a given exit rate rather than its achievability, which is established separately in Section 5. The fallback path includes the key and value repair described in Section 4, so the bookkeeping cost is included.

| Condition | Measured tokens per second | Measured speedup | Estimated speedup |
|---|---|---|---|
| Unmodified model | 7.45 | 1.000 | 1.000 |
| Gated, 26% exit rate (prose) | 8.37 | 1.123 | 1.14 |
| Gated, 52% exit rate (extraction) | 9.68 | 1.298 | 1.31 |
| Gated, 60% exit rate (arithmetic) | 10.09 | 1.353 | 1.38 |

*Table C2. Measured elapsed time against the analytic estimates. Individual repetitions varied by less than two percent within each condition.*

## Appendix D. Complete Results

Tables D1 to D3 give the threshold comparison of Section 5.1 for each model: token-level acceptance at the exit layer, the threshold at which 90 percent of early answers fall within the full model's nucleus, and the resulting exit rate against the exit rate obtained under a single threshold safe for all seven corpora.

| Corpus | Acceptance | Calibrated threshold | Exit rate, global | Exit rate, calibrated | Gain |
|---|---|---|---|---|---|
| Arithmetic | 0.874 | 0.30 | 50.4% | 96.4% | +46.0 |
| Extraction | 0.714 | 0.62 | 30.0% | 66.9% | +36.9 |
| Support | 0.763 | 0.70 | 21.8% | 56.8% | +34.9 |

| Prose | 0.523 | 0.89 | 9.1% | 21.8% | +12.7 |
|---|---|---|---|---|---|
| Python | 0.541 | 0.93 | 18.2% | 29.3% | +11.1 |
| Prose, fixed | 0.543 | 0.91 | 8.4% | 19.2% | +10.8 |
| Chinese | 0.458 | 0.98 | 7.6% | 7.6% | +0.0 |

*Table D1. Qwen2.5-1.5B-Instruct, readout at layer 14 of 28. Global threshold 0.98.*

| **Corpus** | **Acceptance** | **Calibrated threshold** | **Exit rate, global** | **Exit rate, calibrated** | **Gain** |
|---|---|---|---|---|---|
| Arithmetic | 0.870 | 0.39 | 37.2% | 96.0% | +58.8 |
| Extraction | 0.738 | 0.67 | 28.4% | 73.4% | +45.0 |
| Support | 0.764 | 0.80 | 23.7% | 63.0% | +39.3 |
| Python | 0.491 | 0.97 | 16.9% | 24.6% | +7.7 |
| Prose | 0.458 | 0.99 | 8.5% | 8.5% | +0.0 |
| Prose, fixed | 0.416 | 0.99 | 6.5% | 6.5% | +0.0 |
| Chinese | 0.437 | 0.99 | 9.8% | 9.8% | +0.0 |

*Table D2. Qwen2.5-7B-Instruct, readout at layer 14 of 28. Global threshold 0.99.*

| **Corpus** | **Acceptance** | **Calibrated threshold** | **Exit rate, global** | **Exit rate, calibrated** | **Gain** |
|---|---|---|---|---|---|
| Arithmetic | 0.912 | 0.30 | 54.9% | 98.0% | +43.2 |
| Support | 0.783 | 0.65 | 22.0% | 71.9% | +49.9 |
| Extraction | 0.839 | 0.46 | 50.9% | 90.0% | +39.1 |
| Python | 0.639 | 0.89 | 32.7% | 52.5% | +19.8 |
| Prose | 0.532 | 0.97 | 16.8% | 24.1% | +7.3 |
| Prose, fixed | 0.534 | 0.97 | 14.1% | 19.6% | +5.5 |
| Chinese | 0.567 | 0.99 | 15.4% | 15.4% | +0.0 |

*Table D3. Llama-3.1-8B-Instruct, readout at layer 16 of 32. Global threshold 0.99.*

Tables D4 and D5 give closed-loop generation measurements at both thresholds for the two larger models: the fraction of tokens emitted early, the perplexity of the generated text under the unmodified model, the estimated speedup, and the median number of tokens generated before the text falls outside a fixed likelihood bound, which we report as coherent output length.

| **Corpus** | **Exit, $\tau$=0.7** | **Ppl, $\tau$=0.7** | **Coherent, $\tau$=0.7** | **Exit, $\tau$=0.85** | **Ppl, $\tau$=0.85** | **Coherent, $\tau$=0.85** |
|---|---|---|---|---|---|---|
| Arithmetic | 71.8% | 3.55 | 56 | 60.2% | 2.53 | 108 |
| Extraction | 62.1% | 17.34 | 8 | 51.7% | 4.23 | 36 |
| Support | 65.1% | 3.50 | 43 | 50.9% | 2.99 | 54 |
| Python | 54.0% | 5.90 | 44 | 41.6% | 3.70 | 83 |
| Chinese | 38.5% | 6.01 | 32 | 26.8% | 3.64 | 200 |
| Prose | 35.5% | 5.49 | 41 | 26.0% | 2.87 | 98 |
| Prose, fixed | 37.0% | 6.34 | 39 | 24.3% | 3.61 | 120 |

*Table D4. Qwen2.5-7B-Instruct, readout at layer 14 of 28. Unmodified-model perplexity on its own output is 1.33; generations were capped at 200 tokens, so a coherent length of 200 indicates no degradation was detected.*

| **Corpus** | **Exit, $\tau$=0.7** | **Ppl, $\tau$=0.7** | **Coherent, $\tau$=0.7** | **Exit, $\tau$=0.85** | **Ppl, $\tau$=0.85** | **Coherent, $\tau$=0.85** |
|---|---|---|---|---|---|---|
| Arithmetic | 79.3% | 2.07 | 91 | 80.0% | 1.69 | 78 |
| Extraction | 77.0% | 1.86 | 76 | 72.6% | 1.67 | 73 |

| Support | 69.0% | 2.38 | 80 | 54.4% | 1.93 | 84 |
|---|---|---|---|---|---|---|
| Python | 62.9% | 3.56 | 99 | 45.8% | 2.66 | 180 |
| Chinese | 51.7% | 2.89 | 200 | 40.0% | 2.45 | 200 |
| Prose | 49.5% | 3.47 | 105 | 37.6% | 2.33 | 200 |
| Prose, fixed | 46.2% | 3.63 | 110 | 36.5% | 2.47 | 166 |

*Table D5. Llama-3.1-8B-Instruct, readout at layer 16 of 32. Unmodified-model perplexity on its own output is 1.30.*

The arithmetic accuracy measurements are given in Table 5 of Section 9. The comparison of the three softmax-based confidence measures, the reliability of the confidence signal by decile, and the full ten-point threshold sweeps for every model, depth and corpus are in the result files in the repository.